\pdfoutput=1
\def\arxiv{}
\documentclass[letterpaper, 10 pt, conference]{ieeeconf}
\IEEEoverridecommandlockouts
\usepackage{amsmath,amssymb,amsfonts}
\usepackage{xcolor}
\usepackage{graphicx}
\usepackage{stfloats} 
\usepackage[font={small}]{caption}
\usepackage{subcaption}
\usepackage{booktabs}
\usepackage{cite}

\begin{document}

\title{Inertially Significant Humanoid Teleoperation via Whole Body Model Predictive Control and Online Parameter Estimation}

\ifdefined\arxiv
\author{James Foster$^{1,2}$, Beomyeong Park$^{1}$, Luigi Penco$^{1}$, Robert Griffin$^{1,2}$
\thanks{$^{1}$Robotics Lab, Florida Institute for Human \& Machine Cognition, USA}
\thanks{$^{2}$Intelligent Systems \& Robotics, University of West Florida, USA}
\thanks{\texttt{jf127@students.uwf.edu,}}
\thanks{\texttt{\{bpark, lpenco, rgriffin\}@ihmc.org}}}
\else
\author{Author(s) omitted for review
\thanks{Institution(s) omitted for review}
\thanks{\texttt{Email(s) omitted for review}}}
\fi

\maketitle
\thispagestyle{empty}
\pagestyle{empty}

\begin{abstract}
The effective deployment of humanoid robots in practical tasks of interest will require a capability of sensing and adapting to variations in inertial parameters, as occurs during heavy tool use or when carrying significant payload. Such payloads constitute a significant proportion of the combined payload-robot system, the effects of which cannot be ignored by delegating to feedback control. This paper showcases a general solution to this problem by integrating a whole body model predictive controller (WB-MPC) with an extended Kalman filter (EKF) for online estimation of fully physically consistent time-varying inertial parameters. These estimates are fed through to the WB-MPC's full rigid body dynamics model, resulting in more accurate feedforward torques for task demands under external load. Additionally, a virtual reality (VR) interface is provided that enables an operator to provide task-specific references without regard to payload while delegating dynamic balance to the WB-MPC. The framework is validated with simulation and hardware experiments on a humanoid system. 
\end{abstract}


\section{Introduction}

A key promise of humanoid robots lies in performing tasks that involve exerting significant force to manipulated objects, such as lifting a heavy tool or carrying a payload. In many such tasks, the inertia of the manipulated object is a significant fraction of the robot's own.

A promising candidate solution is whole body model predictive control (WB-MPC), which iteratively solves a trajectory optimization over the robot's full rigid body dynamics. It was first demonstrated on a humanoid in~\cite{koenemann2015whole}, and subsequent humanoid implementations have largely targeted locomotion~\cite{dantec2022whole, khazoom2024tailoring, zhang2025whole}. Far fewer works apply WB-MPC to manipulation~\cite{perrot2023step}, and fewer still to objects whose inertia is significant relative to the robot. The closest is~\cite{subburaman2025whole}, which remains in simulation and estimates a modest payload inertia only after task execution. Other model-based approaches either add the load to a centroidal model explicitly~\cite{elobaid2023online}, or absorb it implicitly through direct adaptive control laws bound to a reference model~\cite{sombolestan2023adaptive, minniti2021adaptive}. Reinforcement learning (RL), meanwhile, has produced increasingly robust humanoid locomotion and manipulation~\cite{li2025reinforcement, sun2025ulc}, but typically absorbs load variation in one of two ways. The first is domain randomization~\cite{singh2024robust, dao2024sim}, which trades nominal performance for robustness and tends toward conservative behavior; the second is a learned ``extrinsics'' estimator, whose abstract signals are tightly coupled to the policy that consumes them~\cite{kumar2022adapting}. MPC and RL have recently been combined in~\cite{li2026accelerating}, where a massively parallelized nonlinear centroidal MPC embedded inside policy training yields robust loco-manipulation.

\begin{figure}[t!]
    \centering
    \begin{minipage}[t]{0.49\linewidth}
        \vspace{0pt}
        \includegraphics[width=\linewidth]{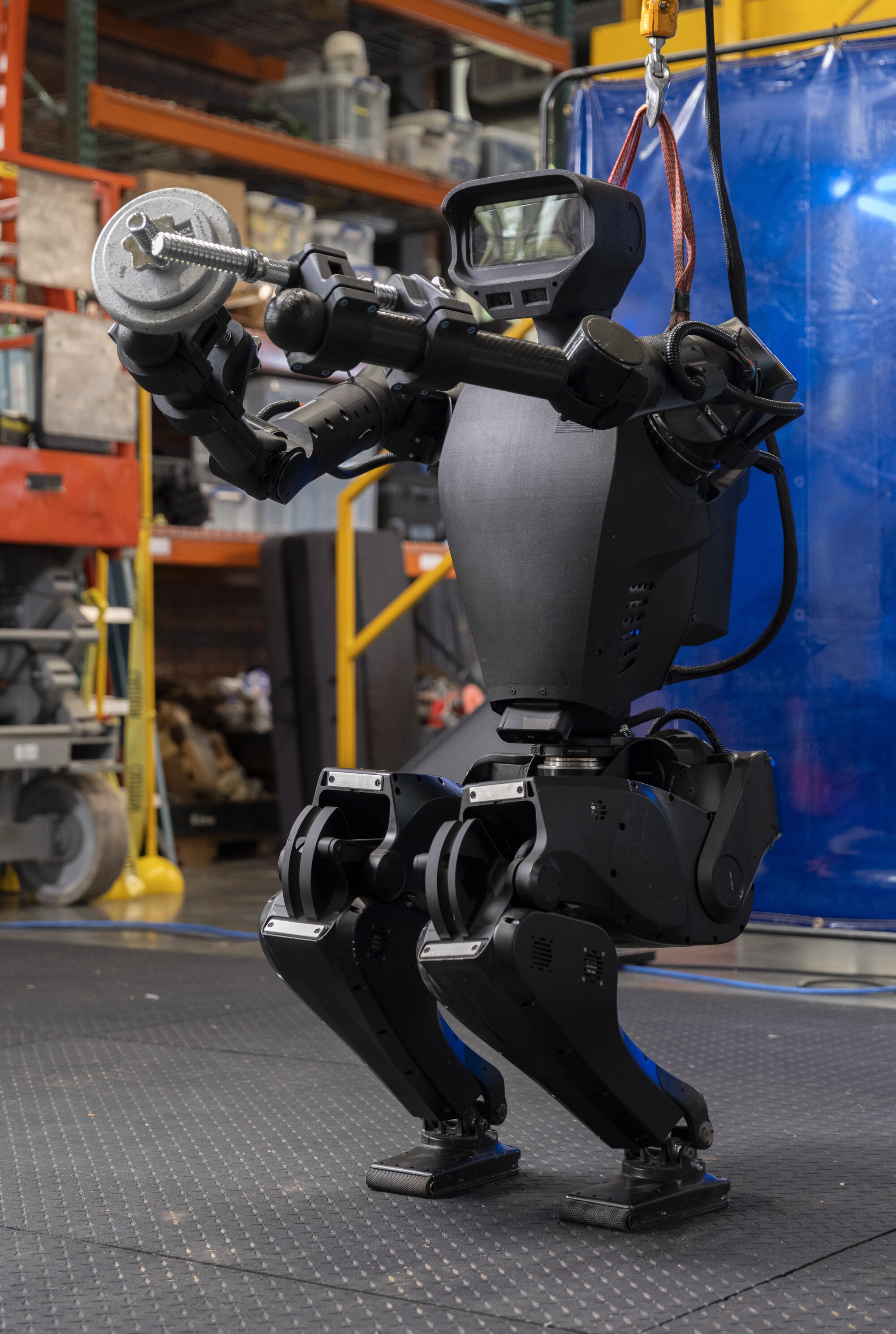}
    \end{minipage}\hfill
    \begin{minipage}[t]{0.49\linewidth}
        \vspace{0pt}%
        \includegraphics[width=\linewidth]{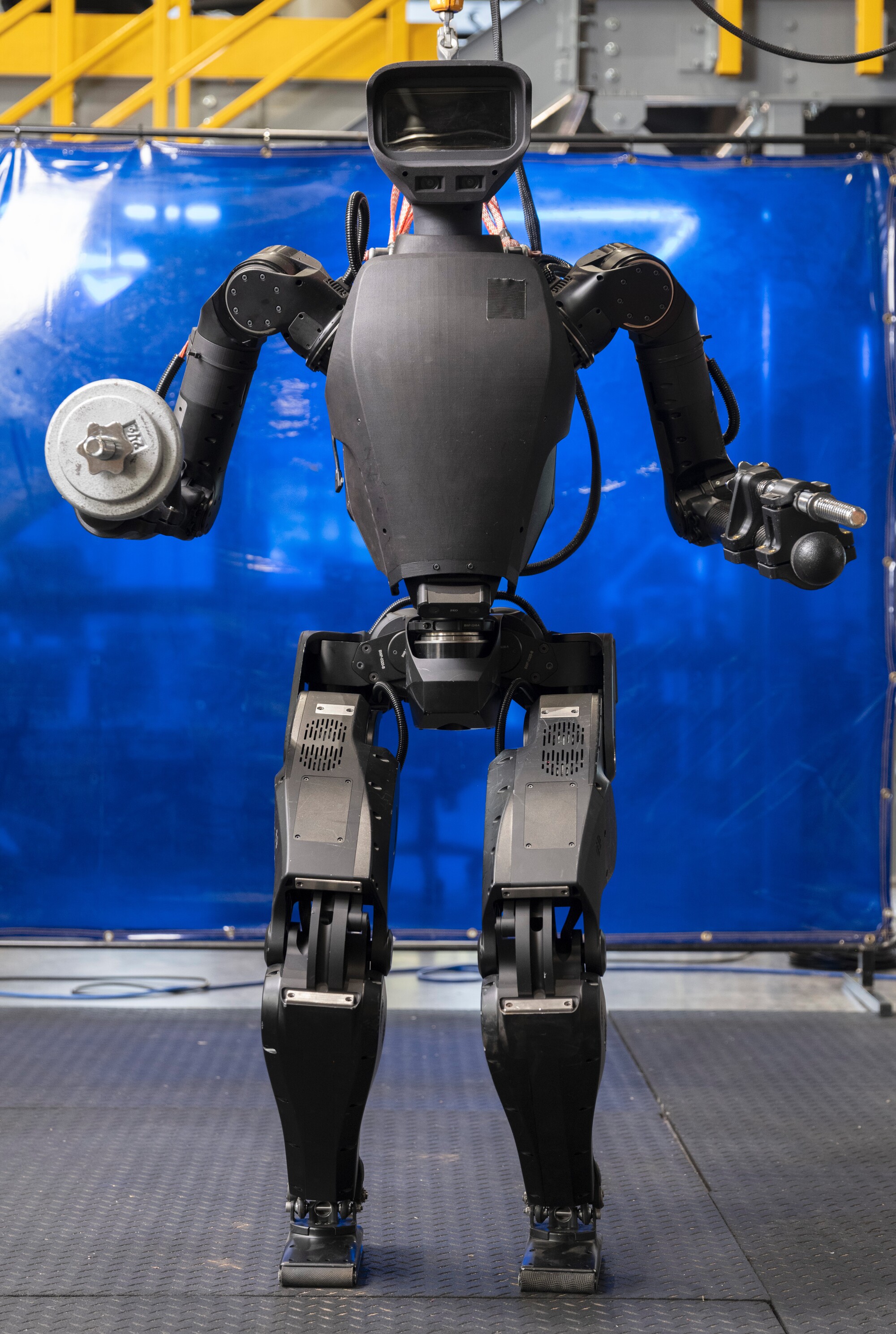}
    \end{minipage}
    \caption{Humanoid teleoperation tasks with a $4.5$\,kg ($10$\,lb) payload fixed to the right forearm link.}
    \label{fig:money}
\end{figure}

A WB-MPC over the full rigid body dynamics can in principle directly command the feedforward effort an added  payload demands, but only insofar as its rigid body dynamics model accurately captures the manipulated inertia~\cite{leboutet2021inertial}. A structure that underpins most identification methods for rigid body inertial parameters is that the inverse dynamics are linear with respect to the inertial parameters~\cite{atkeson1986estimation}. The system identification literature usually focuses on offline identification~\cite{lee2022optimized}; control-oriented online work has instead tended to focus on estimating gross quantities such as the center of mass and momentum~\cite{rotella2015humanoid}, or a payload from previously identified dynamics coefficients~\cite{gaz2017payload} or a momentum observer~\cite{kurdas2022online}. For an estimate to be safely fed into a downstream controller, it must be \textsl{fully physically consistent}~\cite{lee2024robot}, which can be enforced via semidefinite programming~\cite{wensing2017linear}, smooth nonlinear reparameterizations of the inertial parameters~\cite{rucker2022smooth}, or geometric methods~\cite{traversaro2016identification, lee2019geometric}.

Model bias is also problematic when operating the robot via a virtual reality (VR) interface. VR provides operators with an intuitive control interface~\cite{bertrand2024high}, and has become common both as a primary operation method and for data collection of learned control policies~\cite{bostondynamics2025lbm}. With some exceptions, VR operators typically do not have access to force feedback indicating a payload, and so may compensate to mitigate tracking offset, which is burdensome and deleterious to task performance. One such exception is the impressive but complex VR/haptic teleoperation suite in~\cite{baek2025whole}, where online inertial estimates are used to adjust the posture of a wheeled humanoid modeled as a cart-pole system, allowing the manipulation of heavy payloads relative to robot weight. 

In this paper we propose a framework coupling three components to enable intuitive VR teleoperation of humanoids manipulating significant loads via WB-MPC\@. Firstly, a WB-MPC trades off dynamic balance objectives with tight tracking of operator-specified taskspace objectives. Secondly, an extended Kalman filter (EKF) coupled in closed-loop with the WB-MPC estimates fully physically consistent inertial parameters online and passes them to the WB-MPC's rigid body dynamics model, allowing feedforward torque compensation of a priori unknown loads. We include EKF extensions that increase numerical robustness and aid interpretation: we regularize the estimate through a fictitious nominal measurement that keeps the estimate well-conditioned under the weak excitation of teleoperated motion, and estimate deviations from a well-characterized nominal body using a generalized log-Cholesky parameterization~\cite{rucker2022smooth, foster2024physically}. Lastly, a minimal VR teleoperation interface, combined with the aforementioned WB-MPC and EKF, allows an operator to command desired motion without the burden of compensating for added payload, while dynamic balance control is delegated to the WB-MPC\@. We validate the framework on loaded trajectory tracking and balance tasks in simulation, and showcase VR teleoperation under load on hardware for a $23$ degree of freedom humanoid system (Fig.~\ref{fig:money}).

\section{Background}

\begin{figure*}[t!]
    \centering
\includegraphics[width=0.95\textwidth]{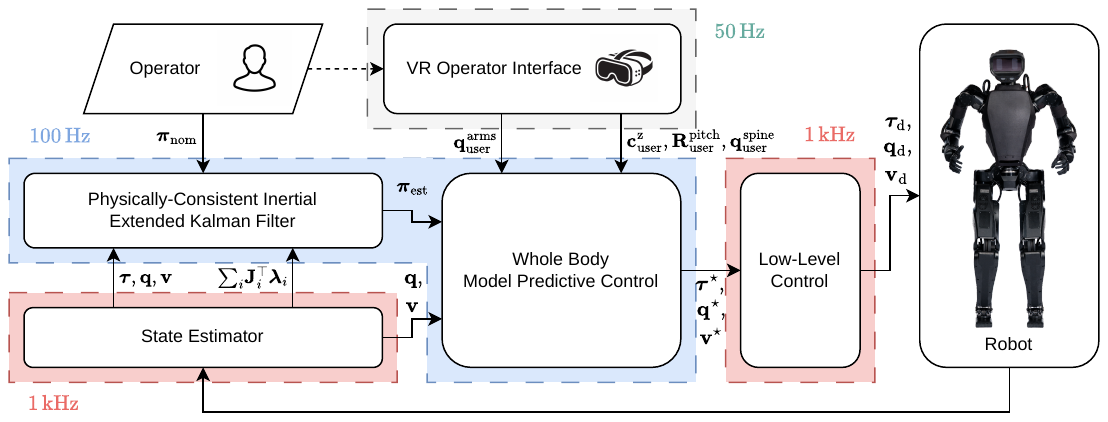}
    \caption{Block diagram of the framework, consisting of the physically consistent inertial EKF (Sec.~\ref{sec:ekf-method}) coupled to the WB-MPC (Sec.~\ref{sec:mpc-method}) and the VR operator interface (Sec.~\ref{sec:vr-method}). The VR operator interface runs on a separate machine.}
    \label{fig:block-diagram}
\end{figure*}

For both the WB-MPC and the EKF for estimating inertial parameters, we start from the assumption that the robot can be adequately modeled by the equations of motion of a floating base rigid body system:
\begin{equation} \label{eq:eom}
\mathbf{M}(\mathbf{q})\mathbf{\dot{v}} + \mathbf{c}(\mathbf{q}, \mathbf{v}) = \mathbf{S}^\top \boldsymbol{\tau} + \sum_{i \in \mathcal{C}}\mathbf{J}_i^\top(\mathbf{q}) \boldsymbol{\lambda}_i,
\end{equation}
where $\mathbf{q}$, $\mathbf{v}$ are the configuration and velocity of the robot, $\boldsymbol{\tau}$ are the joint torques, and $\mathbf{S}$ is the selection matrix for the actuated degrees of freedom. The pair $\mathbf{J}_i$ and $\boldsymbol{\lambda}_i$ are the Jacobian and wrench corresponding to the $i$th contact in the set of external contacts $\mathcal{C}$. $\mathbf{M}$ is the mass matrix and $\mathbf{c}$ collects the centripetal, Coriolis, and gravitational terms.

\subsection{Regressor Formulation of Inverse Dynamics}

There is an equivalent formulation of the inverse dynamics where the inertial parameters enter in a linear fashion~\cite{mistry2009inertial}:
\begin{equation} \label{eq:invdyn-regressor}
\mathbf{Y}(\mathbf{q}, \mathbf{v}, \mathbf{\dot{v}})\boldsymbol{\pi} = \mathbf{S}^\top \boldsymbol{\tau} + \sum_{i \in \mathcal{C}}\mathbf{J}_i^\top(\mathbf{q}) \boldsymbol{\lambda}_i,
\end{equation}
where $\mathbf{Y}$ is a state-dependent regressor matrix that maps the stacked inertial parameters $\boldsymbol{\pi} = [\boldsymbol{\pi}_1^\top, \boldsymbol{\pi}_2^\top, \dots, \boldsymbol{\pi}_N^\top]^\top \in \mathbb{R}^{10N}$ into torque, where $N$ is the number of bodies in the rigid body system. The inertial parameters for the $j^\text{th}$ individual body $\boldsymbol{\pi}_j \in \mathbb{R}^{10}$ consist of its mass $m$, first moment of mass $\boldsymbol{\mu} \in \mathbb{R}^3$, and the unique elements of the symmetric rotational inertia matrix $\mathbf{I} \in \mathbb{R}^{3 \times 3}$:
\begin{equation} \label{eq:pi}
\boldsymbol{\pi}_j =
\begin{bmatrix}
m, & \boldsymbol{\mu}^\top, & \text{vech}(\mathbf{I})^\top
\end{bmatrix}^\top,
\end{equation}
where we drop the $j$ subscript on the components for clarity. The operation $\text{vech}(\cdot)$ vectorizes the lower triangular half of a symmetric matrix.

It is this formulation of the dynamics that forms the backbone of the estimation machinery of the extended Kalman filter (EKF) in Section~\ref{sec:ekf-method}. We use a log-Cholesky parameterization of the inertial parameters~\cite{rucker2022smooth} that ensures the estimates meet the conditions of \textsl{full physical consistency}, which is required in order for a given body's estimate to be physically plausible~\cite{traversaro2016identification}, especially useful as these estimates will be passed to the downstream WB-MPC controller. These conditions are captured compactly by the \textsl{pseudo-inertia}, the symmetric $4 \times 4$ matrix:
\begin{equation} \label{eq:pseudo-inertia}
    \boldsymbol{\Gamma}(\boldsymbol{\pi}_j) = \begin{bmatrix} \boldsymbol{\Sigma} & \boldsymbol{\mu} \\ \boldsymbol{\mu}^\top & m \end{bmatrix},
    \qquad \boldsymbol{\Sigma} = \tfrac{1}{2}\mathrm{Tr}(\mathbf{I})\,\mathbf{1}_3 - \mathbf{I}.
\end{equation}
The pseudo-inertia $\boldsymbol{\Gamma}(\boldsymbol{\pi}_j) \succ 0$ if and only if $\boldsymbol{\pi}_j$ is fully physically consistent~\cite{rucker2022smooth}, and is equivalent to the mass being positive, a positive definite rotational inertia, and a triangle inequality on the principal moments of rotational inertia. 

In the log-Cholesky parameterization of~\cite{rucker2022smooth}, each rigid body is described by $\boldsymbol{\theta}_j \in \mathbb{R}^{10}$:
\begin{equation} \label{eq:theta}
\boldsymbol{\theta}_j = [\alpha, d_1, d_2, d_3, s_{12}, s_{13}, s_{23}, t_1, t_2, t_3]^\top,
\end{equation}
where $\alpha$ scales the density of a unit reference body and $d$, $s$, $t$ apply scaling, shearing, and translation respectively. The parameterization~(\ref{eq:theta}) constructs the pseudo-inertia as the decomposition:
\begin{equation} \label{eq:decomp}
\boldsymbol{\Gamma} = \mathbf{U}\mathbf{U}^\top,
\end{equation}
with the upper-triangular log-Cholesky factor
\begin{equation} \label{eq:log-cholesky}
    \mathbf{U}(\boldsymbol{\theta}_j) = e^{\alpha}
    \begin{bmatrix}
        e^{d_1} & s_{12} & s_{13} & t_1 \\
        0       & e^{d_2} & s_{23} & t_2 \\
        0       & 0       & e^{d_3} & t_3 \\
        0       & 0       & 0       & 1
    \end{bmatrix},
\end{equation}
which keeps $\boldsymbol{\Gamma}$ positive definite for \emph{any} $\boldsymbol{\theta}_j$, so every estimate is physically consistent by construction. A smooth, invertible map $\boldsymbol{\pi}_j = \mathbf{g}(\boldsymbol{\theta}_j)$ with closed-form Jacobian makes it simple to switch between the two parameterizations.

\subsection{Whole Body Model Predictive Control}
\label{sec:wbmpc-background}

Defining the state $\mathbf{x} = (\mathbf{q}, \mathbf{v})$ and the control $\mathbf{u} = \boldsymbol{\tau}$, our WB-MPC repeatedly solves, at each control instant and from the current estimated state $\hat{\mathbf{x}}$, the discrete-time optimal control problem (OCP)
\begin{subequations} \label{eq:ocp}
\begin{align}
    \min_{\mathbf{x}_{0:K}, \, \mathbf{u}_{0:K-1}} \quad
        & \ell_K(\mathbf{x}_K) + \sum_{k=0}^{K-1} \ell_k(\mathbf{x}_k, \mathbf{u}_k) \label{eq:ocp-cost} \\
    \text{s.t.} \quad
        & \mathbf{x}_0 = \hat{\mathbf{x}}, \label{eq:ocp-init} \\
        & \mathbf{x}_{k+1} = \mathbf{f}_\text{dyn}(\mathbf{x}_k, \mathbf{u}_k; \boldsymbol{\pi}), \label{eq:ocp-dyn} \\
        & \mathbf{f}_\text{constr}(\mathbf{x}_k, \mathbf{u}_k) \le \mathbf{0}, \label{eq:ocp-constr}
\end{align}
\end{subequations}
over a horizon of $K$ stages, where $\ell_k$ and $\ell_K$ are the running and terminal costs and $\mathbf{f}_\text{constr}$ collects the path constraints.

The dynamics model is obtained from by manipulating the equations of motion (\ref{eq:eom}) to obtain the forward dynamics:
\begin{equation} \label{eq:forward-dynamics}
    \dot{\mathbf{v}} = \mathbf{M}(\mathbf{q})^{-1}\!\left( \mathbf{S}^\top \boldsymbol{\tau} + \sum_{i \in \mathcal{C}} \mathbf{J}_i^\top(\mathbf{q})\,\boldsymbol{\lambda}_i - \mathbf{c}(\mathbf{q}, \mathbf{v}) \right),
\end{equation}
which can be integrated to yield the discrete transition $\mathbf{x}_{k+1} = \mathbf{f}_\text{dyn}(\mathbf{x}_k, \mathbf{u}_k; \boldsymbol{\pi})$ of (\ref{eq:ocp-dyn}). Because $\mathbf{M}$ and $\mathbf{c}$ depend on the inertial parameters $\boldsymbol{\pi}$, it is through $\mathbf{f}_\text{dyn}$ that the online estimates of these parameters enter the controller.
\section{Method}

A block diagram of our approach is shown in Fig.~\ref{fig:block-diagram}. Our contributions center around three components: an EKF that estimates fully physically consistent inertial parameters online (Section~\ref{sec:ekf-method}), a WB-MPC that plans against a full rigid body dynamics updated by the EKF (Section~\ref{sec:mpc-method}), and a VR teleoperation interface that supplies task references to the WB-MPC (Section~\ref{sec:vr-method}).

\subsection{Extended Kalman Filter}
\label{sec:ekf-method}

Our estimator adopts the physically consistent inertial EKF formulation of~\cite{foster2024physically}; we summarize it below, then introduce two extensions that improve robustness under teleoperated motion. We assume a prior offline system identification resolves the majority of the robot's inertial parameters, and that the task dictates which links vary inertially, so the regressor in (\ref{eq:invdyn-regressor}) can be split into an estimated and a nominal contribution $\mathbf{Y}\boldsymbol{\pi} = \mathbf{Y}^\text{est}\boldsymbol{\pi}^\text{est} + \mathbf{Y}^\text{nom} \boldsymbol{\pi}^\text{nom}$. The latter nominal contribution can be resolved by a single inverse-dynamics call $\boldsymbol{\tau}^\text{nom}$ with the to-be estimated parameters zeroed, giving
\begin{equation} \label{eq:decomp-tau}
    \mathbf{Y}\boldsymbol{\pi} = \mathbf{Y}^\text{est}\boldsymbol{\pi}^\text{est} + \boldsymbol{\tau}^\text{nom}.
\end{equation}
The measurement model is formed by isolating the estimated contribution when (\ref{eq:decomp-tau}) is substituted into (\ref{eq:invdyn-regressor}): replacing $\mathbf{S}^\top\boldsymbol{\tau}$ with the observed joint torques $\boldsymbol{\tau}^\text{obs}$ and the measured contact wrenches $\boldsymbol{\lambda}_i$ yields the residual torque
\begin{equation} \label{eq:delta-tau}
    \Delta \boldsymbol{\tau}_k = \boldsymbol{\tau}^\text{obs} - \boldsymbol{\tau}^\text{nom} + \sum_{i \in \mathcal{C}} \mathbf{J}^\top_i \boldsymbol{\lambda}_i \approx \mathbf{Y}_k^\text{est} \boldsymbol{\pi}^\text{est}_k.
\end{equation}

By forfeiting the hitherto linear parameterization of $\boldsymbol{\pi}$ in (\ref{eq:pi}) and moving to the parameterization of $\boldsymbol{\theta}$ in (\ref{eq:theta}), a fully physically consistent estimate is guaranteed from the now nonlinear measurement model $\mathbf{h}$ at every iteration, enforcing the use of an EKF as opposed to a regular Kalman filter:
\begin{equation} \label{eq:measurement-lc}
    \Delta \boldsymbol{\tau}_k = \underbrace{\mathbf{Y}^\text{est}_k\,\mathbf{g}(\boldsymbol{\theta}^\text{est}_k)}_{\mathbf{h}(\boldsymbol{\theta}^\text{est}_k)} + \boldsymbol{\nu}_k, \quad \boldsymbol{\nu}_k \sim \mathcal{N}(\mathbf{0}, \mathbf{R}_k),
\end{equation}
abusing notation by reusing $\boldsymbol{\nu}$ in a different parameterization. The invertible mapping between parameterizations $\boldsymbol{\pi}^\text{est} = \mathbf{g}(\boldsymbol{\theta}^\text{est})$~\cite{rucker2022smooth} and its closed-form Jacobian, denoted $\mathbf{G}$, are used to form the measurement Jacobian $\mathbf{H}$ by applying the chain rule: $\mathbf{H}_k = \frac{\partial \mathbf{h}}{\partial \boldsymbol{\theta}} = \mathbf{Y}^\text{est}_k \mathbf{G}_k$.

The process model is assumed constant, driven by noise:
\begin{equation}
\boldsymbol{\theta}_k^\text{est} = \underbrace{\boldsymbol{\theta}_{k-1}^\text{est}}_{\mathbf{f}(\boldsymbol{\theta}_{k-1}^\text{est})} + \boldsymbol{\eta}_k, \quad \boldsymbol{\eta}_k \sim \mathcal{N}(\mathbf{0}, \mathbf{Q}_k),
\end{equation}
so the process Jacobian is the identity matrix $\mathbf{F}_k = \frac{\partial \mathbf{f}}{\partial \boldsymbol{\theta}} =\mathbf{I}$.

Below we describe two extensions to this framework which greatly assisted teleoperation tasks.

\subsubsection{Estimating deviations from a nominal body} \label{sec:deviations}
Reference work~\cite{foster2024physically} estimated a given link's parameters ``from scratch'', as can be noted by seeing that (\ref{eq:log-cholesky}) transforms a unit reference body, implicit in (\ref{eq:decomp}) as $\boldsymbol{\Gamma} = \mathbf{U} \mathbf{I}_{4 \times 4} \mathbf{U}^\top$. In our implementation, we found that to cause tuning issues when links of vastly different size are included in the same estimation, as is common with humanoids. Instead, we estimate a physically consistent \textsl{deviation} from a known nominal inertia using the \textsl{generalized} log-Cholesky parameterization of~\cite{rucker2022smooth}. To anchor at the nominal body instead, we replace that unit reference with the nominal pseudo-inertia $\boldsymbol{\Gamma}(\boldsymbol{\pi}_j^\text{nom})$,
\begin{equation} \label{eq:gen-log-cholesky}
    \boldsymbol{\Gamma}(\boldsymbol{\theta}_j^\text{est})
    = \mathbf{U}(\boldsymbol{\theta}_j^\text{est})\,
      \boldsymbol{\Gamma}(\boldsymbol{\pi}_j^\text{nom})\,
      \mathbf{U}(\boldsymbol{\theta}_j^\text{est})^\top,
\end{equation}
and the standard parameters follow from $\boldsymbol{\Gamma}(\boldsymbol{\theta}_j^\text{est})$ by the same smooth, invertible map $\mathbf{g}$ and closed-form Jacobian $\mathbf{G}$ as before, but now applied to the composition of the estimated $\boldsymbol{\theta}^\text{est}$ and nominal $\boldsymbol{\theta}^\text{nom}$ log-Cholesky parameter sets:
\begin{equation} \label{eq:compose}
\boldsymbol{\pi}_j^\text{est} = \mathbf{g}(\boldsymbol{\theta}_j^\text{est} \oplus \boldsymbol{\theta}_j^\text{nom}),
\end{equation}
where $\boldsymbol{\theta}_j^\text{nom} = \mathbf{g}^{-1}(\boldsymbol{\pi}_j^\text{nom})$ and $\oplus$ is the composition defined by $\mathbf{U}(\boldsymbol{\theta}_j^\text{est} \oplus \boldsymbol{\theta}_j^\text{nom}) = \mathbf{U}(\boldsymbol{\theta}_j^\text{est})\,\mathbf{U}(\boldsymbol{\theta}_j^\text{nom})$. Setting $\boldsymbol{\theta}_j^\text{est} = \mathbf{0}$ reproduces the nominal body, and the entries of $\boldsymbol{\theta}_j^\text{est}$ now measure the change in density, scale, shear and translation of the estimated body from the nominal one. We have found this to be more numerically robust, and a more natural way of describing deviations from a nominal inertia due to payload.

\subsubsection{Regularizing weakly-observable parameters} \label{sec:weakly-observable}
If there are weakly-observable parameters due to lack of excitation~\cite{lee2024robot}, we can append a fictitious measurement that the parameter deviation is zero (the nominal body), $\mathbf{0} = \boldsymbol{\theta}_k^\text{est} + \boldsymbol{\nu}_k^\text{reg}$ with $\boldsymbol{\nu}_k^\text{reg} \sim \mathcal{N}(\mathbf{0}, \mathbf{R}^\text{reg})$, with $\mathbf{R}^\text{reg}$ weighted only on channels we wish to regularize. Stacked beneath the torque measurement (\ref{eq:measurement-lc}), it augments the EKF update as
\begin{equation} \label{eq:tikhonov}
    \begin{bmatrix} \Delta\boldsymbol{\tau}_k \\ \mathbf{0} \end{bmatrix}
    = \begin{bmatrix} \mathbf{h}(\boldsymbol{\theta}_k^\text{est}) \\
                      \boldsymbol{\theta}_k^\text{est} \end{bmatrix}
    + \begin{bmatrix} \boldsymbol{\nu}_k \\ \boldsymbol{\nu}_k^\text{reg} \end{bmatrix},
\end{equation}
with the augmented measurement Jacobian and covariance:
\begin{equation}
    \quad
    \mathbf{H}_k^\text{aug} = \begin{bmatrix} \mathbf{H}_k \\ \mathbf{I} \end{bmatrix},
    \quad
    \mathbf{R}_k^\text{aug} = \begin{bmatrix} \mathbf{R}_k & \\ & \mathbf{R}^\text{reg} \end{bmatrix}.
\end{equation}
This is equivalent to a maximum a posteriori update with a Gaussian prior centered at the nominal body, applied only to the weakly-observable channels. For example, in Section~\ref{sec:simulation}, we found that mass and center of mass of the loaded forearm links are well-excited under teleoperated motion, but rotational inertia is excited weakly and drifts if left free. 

\subsection{Whole Body Model Predictive Control}
\label{sec:mpc-method}

\begin{table*}[t]
\centering
\caption{Cost terms comprising the WB-MPC\@. Without input, desired quantities $*_\text{des}$ are the ``Standard reference'' column.}
\label{tab:costs}
\footnotesize
\setlength{\tabcolsep}{6pt}
\begin{tabular}{@{}lllll@{}}
\toprule
Cost & Residual $\mathbf{r}$ & Type & Activation & Standard reference \\
\midrule
Upper body jointspace tracking         & $\mathbf{K}_d(\mathbf{v}^\text{upper}-\mathbf{v}^\text{upper}_\text{des}) + \mathbf{K}_p(\mathbf{q}^\text{upper}-\mathbf{q}^\text{upper}_\text{des})$ & $2^\text{nd}$ order & Weighted Quad.\ & Home config., zero velocity \\
Center of mass tracking                  & $\mathbf{K}_d(\dot{\mathbf{c}}-\dot{\mathbf{c}}_\text{des}) + \mathbf{K}_p(\mathbf{c}-\mathbf{c}_\text{des})$ & $2^\text{nd}$ order & Weighted Quad.\ & $(0, 0, \mathbf{c}_\text{home}^z)$ \\
Pelvis orientation            & $\mathbf{K}_d(\boldsymbol{\omega}-\boldsymbol{\omega}_\text{des}) + \mathbf{K}_p\log(\mathbf{R}_\text{des}^\top\mathbf{R})$ & $2^\text{nd}$ order & Weighted Quad.\ & $(0, 0, 0, 1)$ (quaternion) \\
Foot placement                & $\mathbf{K}_d(\dot{\mathbf{p}}-\dot{\mathbf{p}}_\text{des}) + \mathbf{K}_p(\mathbf{p}-\mathbf{p}_\text{des})$ & $2^\text{nd}$ order & Weighted Quad.\ & Home config. equivalent \\
Angular momentum           & $(\dot{\mathbf{l}}-\dot{\mathbf{l}}_\text{des}) + \boldsymbol{\alpha}(\mathbf{l}-\mathbf{l}_\text{des})$ & $1^\text{st}$ order & Weighted Quad.\ & $\mathbf{0}$ \\
Lower body state regularization & $\mathbf{x}^\text{legs}\ominus\mathbf{x}^\text{legs}_\text{des}$ & Standard & Weighted Quad.\ & Home config., zero velocity \\
Acceleration regularization\      & $\dot{\mathbf{v}}$ & Standard & Weighted Quad.\ & $\mathbf{0}$ \\
Torque regularization         & $\boldsymbol{\tau}$ & Standard & Weighted Quad.\ & $\mathbf{0}$ \\
Wrench regularization\          & $\boldsymbol{\lambda}_i-\boldsymbol{\lambda}_{i,\text{des}}$ & Standard & Weighted Quad.\ & $(0, 0, \tfrac{mg}{\text{dim}(\mathcal{C})}, 0, 0, 0)$ \\
Center of pressure            & $\mathbf{A}_i^\text{cop}\,\boldsymbol{\lambda}_i$ & Standard & Weighted Quad.\ & Foot center \\
Wrench cone            & $\mathbf{A}_i^\text{wc}\,\boldsymbol{\lambda}_i$ & Constraint & Quad.\ Barrier & Wrench cone, $25$\,N margin  \\
\bottomrule
\end{tabular}
\end{table*}

We solve the OCP~(\ref{eq:ocp}) with the feasibility-driven differential dynamic programming (FDDP) solver of \texttt{crocoddyl}~\cite{mastalli2020crocoddyl} which, through its internal use of \texttt{pinocchio}~\cite{carpentier2019pinocchio}, furnishes the forward dynamics~(\ref{eq:forward-dynamics}) and their analytical derivatives at each stage.

The running cost $\ell_k$ of~(\ref{eq:ocp}) is mostly a weighted sum of least-squares terms: for a generic cost residual $\mathbf{r}$ we use a weighted quadratic $\lVert\mathbf{r}\rVert_{\mathbf{W}}^2$ with positive-definite weight $\mathbf{W}$. Writing $\mathbf{e}$ for a generic task error of a tracked quantity minus its reference, a regularization task penalizes the error directly through the standard residual
\begin{equation} \label{eq:res-std}
    \mathbf{r}^\text{std} = \mathbf{e}.
\end{equation}
Inspired by and similar to~\cite{ubbink2024instantaneous}, we have had success with residuals that penalize deviation from a \textsl{desired response}, be that first or second order:
\begin{align}
    \mathbf{r}^\text{1st} &= \dot{\mathbf{e}} + \boldsymbol{\alpha} \mathbf{e}, \label{eq:conv-residual-1st} \\
    \mathbf{r}^\text{2nd} &= \ddot{\mathbf{e}} + \mathbf{K}_d\,\dot{\mathbf{e}} + \mathbf{K}_p\,\mathbf{e}, \label{eq:conv-residual-2nd}
\end{align}
where $\boldsymbol{\alpha}$, $\mathbf{K}_p$, and $\mathbf{K_d}$ are appropriate gains for the task. An in-depth study of the effect of these \textsl{response-based} residuals for humanoid locomotion on the WB-MPC solution is underway. In agreement with~\cite{ubbink2024instantaneous}, we note that these residuals make tuning easier by decoupling task priority from desired closed-loop bandwidth, leading to more horizon-invariant behavior. These characteristics are particularly useful for tracking the user references from VR, which are uniform over the WB-MPC prediction horizon.

The cost terms used in the WB-MPC are collected in Table~\ref{tab:costs}, where we partition the configuration by kinematic sub-chain $\mathbf{q} = (\mathbf{q}^\text{base}, \mathbf{q}^\text{legs}, \mathbf{q}^\text{spine}, \mathbf{q}^\text{arms})$, so that each cost can be appropriately scoped to the sub-chains on which it acts. We also define the collection $\mathbf{q}^\text{upper} = (\mathbf{q}^\text{spine}, \mathbf{q}^\text{arms})$ for convenience. Differences on the state manifold are written $\mathbf{x} \ominus \mathbf{x}_\text{des}$, reducing to subtraction on the joint variables and to the $SE(3)$ logarithm on the floating base. In the special case of wrench cone constraints, as FDDP cannot natively path constraints in~(\ref{eq:ocp-constr}), they are absorbed into the cost as soft penalties through a quadratic barrier activation.

We employ no specific alterations to the terminal cost $\ell_K$. We find that the combination of standard regularization costs, coupled with the response-based costs on user VR input, suffice to induce horizon-invariant, stable behavior.

\subsection{Virtual Reality Teleoperation}
\label{sec:vr-method}

In order to stream references to the WB-MPC, the VR operator's tracked device poses consisting of the two joysticks and the headset are resolved into jointspace targets by an inverse kinematics (IK) routine running on a separate operator-side computer. The WB-MPC consumes only the upper-body portion of this solution: the IK routine's output configuration and velocity for the arms become the reference $(\mathbf{q}^\text{arms}_\text{user}, \mathbf{v}^\text{arms}_\text{user})$ of the jointspace tracking cost of Section~\ref{sec:mpc-method}.

Although a user study is outside the scope of this work, we specify compact VR joystick controller bindings for streaming pelvis pitch $\mathbf{R}^\text{pitch}_\text{user}$, center of mass (CoM) height $\mathbf{c}^z_\text{user}$, and spine yaw $\mathbf{q}^\text{spine}_\text{user}$ references to the WB-MPC in an attempt to ease the VR operator burden. The set of quantities streamed from the VR interface to the WB-MPC is given in Table~\ref{tab:vr-mapping}. In contrast to the whole-body VR/haptic interface in~\cite{baek2025whole}, we conjecture that a simple joystick-based interface makes operation more intuitive; the operator commands the base pose of a bi-manual manipulation system via the joystick, and then teleoperates the arms.

\begin{table}[t]
\centering
\caption{VR inputs for streaming to WB-MPC.}
\label{tab:vr-mapping}
\footnotesize
\setlength{\tabcolsep}{4pt}
\begin{tabular}{@{}p{0.42\columnwidth} p{0.46\columnwidth}@{}}
\toprule
WB-MPC cost term & VR operator input \\
\midrule
Arm jointspace $\mathbf{q}^\text{arms}_\text{user}$   & Handheld controllers \\
Spine jointspace $\mathbf{q}^\text{spine}_\text{user}$  & Right thumbstick, horizontal \\
Pelvis pitch $\mathbf{R}^\text{pitch}_\text{user}$ & Right thumbstick, vertical \\
Center of mass height $\mathbf{c}^\text{z}_\text{user}$ & Left thumbstick, vertical \\
\bottomrule
\end{tabular}
\end{table}

\subsection{Practical Implementation Details}

The framework in Fig.~\ref{fig:block-diagram} is implemented on two machines. The WB-MPC, EKF, and state estimator are deployed inside a single real-time process on an onboard control computer. The state estimator and low-level control run on a $1$\,kHz thread; the EKF and WB-MPC run on a $100$\,Hz thread. The VR kinematic streaming toolbox of Section~\ref{sec:vr-method} runs on a separate operator desktop computer and streams its references via ROS2 messages at a best-effort rate of $50$\,Hz, decoupling the operator interface from the real-time loop.

The inertial EKF reads $\mathbf{q}$, $\mathbf{v}$, and $\boldsymbol{\tau}^\text{obs}$ from the state estimator and the contact wrenches $\boldsymbol{\lambda}_i$ are computed from the leg torques, $\boldsymbol{\tau}_{\text{leg},i} = \mathbf{J}_i^\top \boldsymbol{\lambda}_i$. The joint accelerations $\dot{\mathbf{v}}$ required by the measurement model are reconstructed by a low-pass-filtered finite difference of $\mathbf{v}$. Each tick, the estimate is mapped back to $\boldsymbol{\pi}^\text{est}$, rate-limited to prevent abrupt jumps, and written to the WB-MPC's dynamics model.

The WB-MPC horizon length $K = 30$ nodes with an Euler discretization of $\Delta t = 20$\,ms which resolves to a prediction horizon of $0.6$\,s. The result of a WB-MPC solve is a tape of feedforward torques and states: $\boldsymbol{\tau}^\star = [\boldsymbol{\tau}^\star_0, \dots, \boldsymbol{\tau}^\star_{K-1}]$, $\mathbf{q}^\star = [\mathbf{q}^\star_0, \dots, \mathbf{q}^\star_K]$, and $\mathbf{v}^\star = [\mathbf{v}^\star_0, \dots, \mathbf{v}^\star_K]$. The low-level controller closes the loop on the slower WB-MPC solution by interpolating cubic Hermite splines through the planned $\mathbf{q}^\star$ and $\mathbf{v}^\star$ setpoints and tracking them with a PD law that supplements the feedforward torque from the WB-MPC\@.
\section{Results}
\label{sec:results}

To verify the framework, we test on a humanoid system with $23$ internal degrees of freedom: six per leg, one spine yaw, four per arm, and two in the neck (the neck is not considered in the EKF/WB-MPC scheme, it is PD-controlled around a default configuration). All simulation and hardware experiments use the EKF configuration given in Table~\ref{tab:ekf-config}.

\begin{table}[tb]
\centering
\caption{EKF configuration used for experiments. The mass ($\alpha$) and center of mass ($t$) channels are unregularized (---); only the rotational inertia channels ($d$, $s$) carry a prior.}
\label{tab:ekf-config}
\footnotesize
\begin{tabular}{@{}ll@{}}
\toprule
Parameter & Value \\
\midrule
Process $\mathbf{Q}$: $\alpha$ & $10^{-3}$ \\
Process $\mathbf{Q}$: $d$, $s$, $t$ & $10^{-2}$ \\
Prior $\mathbf{R}^\text{reg}$: $d$, $s$ & $10^{-2}$ \\
Prior $\mathbf{R}^\text{reg}$: $\alpha$, $t$ & --- \\
Measurement $\mathbf{R}$: floating base, spine & $10^{3}$ \\
Measurement $\mathbf{R}$: arms, legs & $10^{2}$ \\
\bottomrule
\end{tabular}
\end{table}

\subsection{Simulation} \label{sec:simulation}

\subsubsection{Pre-recorded trajectory tracking with/without EKF} \label{sec:sim-wash}

\begin{figure}[tb]
    \centering
    \includegraphics[width=\linewidth,height=!]{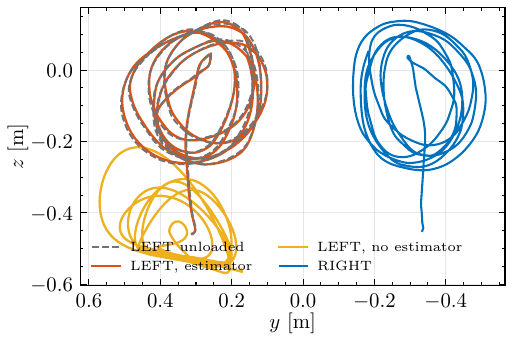}
    \caption{Forearm taskspace paths (chest frame, frontal $y$--$z$ plane) for pre-recorded trajectory.}
    \label{fig:wash-taskspace}
\end{figure}

\begin{figure}[tb]
    \centering
    \includegraphics[width=\linewidth,height=!]{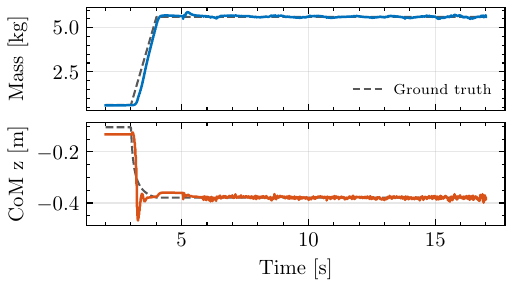}
    \caption{Estimated forearm mass and CoM vs. ground truth.}
    \label{fig:wash-estimate}
\end{figure}

To isolate the estimator's effect on tracking from operator variability, we drive the robot with a pre-recorded teleoperation trajectory, consisting of a window-washing motion that sweeps both forearms across the frontal plane. Though pre-recorded, it is fed through such that the WB-MPC cannot anticipate the trajectory. At $3$\,s, the left forearm (nominal mass $0.62$\,kg) has a $5$\,kg payload mounted at the forearm tip ($0.41$\,m down from the elbow joint), while the right is unloaded and serves as a within-trial reference. We compare taskspace tracking of the loaded forearm with the EKF disabled versus enabled, feeding estimates to the WB-MPC's dynamics model in Fig.~\ref{fig:wash-taskspace} and Table~\ref{tab:wash-metrics}. With no adaptation, the loaded left forearm sags significantly from its unloaded path, accumulating a vertical taskspace root mean square error (RMSE) of $389.2$\,mm, an over $25\times$ increase compared to the error with adaptation. The horizontal taskspace error is less severe, though still significant; without adaptation the RMSE is $67.8$\,mm while with adaptation it shrinks over $10\times$ to $6.4$\,mm. The estimated mass and center of mass track the added payload within two seconds of application, (Fig.~\ref{fig:wash-estimate}).

Table~\ref{tab:ablation} shows the value of estimating deviations from a known nominal body (Section~\ref{sec:deviations}) and regularizing weakly-observable parameters (Section~\ref{sec:weakly-observable}). Only the mass ($\alpha$) and center of mass down the forearm ($t_3$) have any significant deviation from nominal. Mapping these deviations through the generalized log-Cholesky parameterization recovers the payload: the mass channel $\alpha\approx1.10$ scales the nominal by $e^{2\alpha}\approx9.1$, taking the $0.62$\,kg forearm to $5.6$\,kg, while $t_3\approx-0.28$ moves the center of mass from the nominal $-0.10$\,m down the link to $-0.38$\,m. Regarding adding a fictitious measurement, we see that the mass and center of mass are recovered regardless. The difference is the weakly-observable rotational inertia channels ($d$, $s$). If left free they wander from nominal, whereas the fictitious measurement of inertia induces a prior that holds them near zero, motivating the EKF configuration in Table~\ref{tab:ekf-config} for less ``slosh'' in estimates~\cite{foster2024physically}.

\begin{table}[tb]
\centering
\caption{Taskspace error statistics (frontal $y$--$z$ plane) for pre-recorded trajectory, loaded vs. unloaded baseline.}
\label{tab:wash-metrics}
\footnotesize
\begin{tabular}{@{}lcc@{}}
\toprule
Condition & RMSE $y$ [mm] & RMSE $z$ [mm] \\
\midrule
No estimator   & 67.8 & 389.2 \\
With estimator &  6.4 &  14.2 \\
\bottomrule
\end{tabular}
\end{table}

\begin{table}[tb]
\centering
\caption{Effects of parameter prior. Each parameter group entry is the RMS of the Euclidean norm of that group's deviation from nominal: $|\alpha|$, $||d||_2$, $||s||_2$, $||t||_2$.}
\label{tab:ablation}
\footnotesize
\begin{tabular}{@{}lccccc@{}}
\toprule
 & Elbow RMSE & \multicolumn{4}{c}{RMS deviation from nominal} \\
\cmidrule(l){3-6}
Prior regime & [deg] & $\alpha$ & $d$ & $s$ & $t$ \\
\midrule
No prior            & 1.27 & 1.11 & 0.11 & 0.09 & 0.28 \\
Inertia-only        & 1.27 & 1.11 & $4{\times}10^{-5}$ & $1{\times}10^{-5}$ & 0.28 \\
\bottomrule
\end{tabular}
\end{table}


\subsubsection{Posture adjustment under load with/without EKF}

\begin{figure}[tb]
    \centering
    \includegraphics[width=\linewidth,height=!]{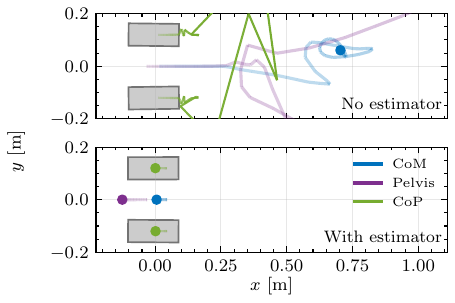}
    \caption{Overhead ($x$--$y$ plane) view of robot balance quantities during static hold.}
    \label{fig:hold-overhead}
\end{figure}

The second experiment stresses whole-body balance of the WB-MPC, rather than arm tracking. The robot is commanded to extend both arms horizontally in front of the torso to replicate a two-handed load carrying posture. Each forearm is then loaded with $5$\,kg ($10$\,kg in total) mounted at the forearm tips, ramped up linearly over two seconds once the arms have reached the forward-hold pose. With the EKF disabled, the center of mass and pelvis drift steadily forward until the centers of pressure (CoP) of the feet leave the support polygon and the robot tips unstable (Fig.~\ref{fig:hold-overhead}, top). With the EKF enabled the controller accounts for the added mass by modifying its internal posture: driving the pelvis backward in order to maintain the center of mass within the support polygon and its pelvis upright (Fig.~\ref{fig:hold-overhead}, bottom). This counter-balance is emergent from the WB-MPC costs in Table~\ref{tab:costs}.

\subsection{Hardware} \label{sec:hardware}

For the hardware experiments, we mount adjustable dumbbells that extend axially past the end of the link; as we do not have access to heavy duty grippers, we ``grasp'' a payload by manually mounting weights at runtime. Due to recurring hardware issues on the left arm, we only load the right arm and take this as an opportunity to show how the WB-MPC/EKF can handle challenging asymmetric loads.

\subsubsection{Runtime loading}

\begin{figure}[t]
    \centering
    \includegraphics[width=\linewidth]{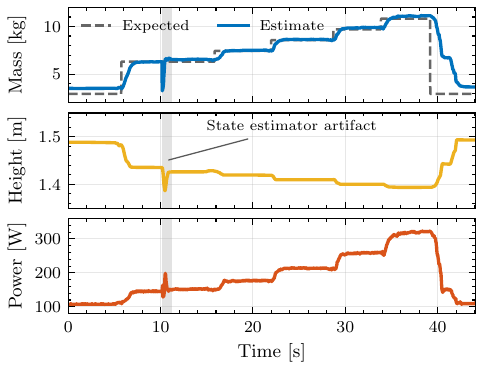}
    \caption{Mass estimate, end-effector taskspace height, and whole-robot electrical power draw for the incremental hardware loading of the right forearm held horizontally in front of the robot. The shaded interval marks a state estimator artifact.}
    \label{fig:loading-estimate}
\end{figure}

With both arms horizontally extended in front of the torso, we incrementally mount weight plates onto the right forearm cuff at runtime: first a $3.4$\,kg ($7.5$\,lb) plate, then four $1.1$\,kg ($2.5$\,lb) plates one at a time, before removing the entire load. The estimated forearm mass tracks the ground-truth loading staircase closely (Fig. \ref{fig:loading-estimate}, top): relative to the pre-load estimate, the filter recovers $89\%$ of the initial $3.4$\,kg plate and, as further plates are added, $88$, $91$, $94$, and $96\%$ of the cumulative load at $4.5$, $5.7$, $6.8$, and $7.9$\,kg ($10$, $12.5$, $15$, and $17.5$\,lb). The estimate carries a small positive bias ($0.6$\,kg) relative to the CAD nominal, indicating some model mismatch effects.

The height of the forearm end-effector sags with each added plate, dropping $94$\,mm from its unloaded to fully loaded height (Fig. \ref{fig:loading-estimate}, middle). Though modest compared to the complete failure that results without closed-loop estimation, we believe this leaves scope for more thorough identification of joint friction/damping parameters; once the load is removed the tip returns to within $2$\,mm of its original height. The robot's total electrical power draw, measured at the power-management board, rises from $108$\,W unloaded to $318$\,W at peak load, a $2.9\times$ increase (Fig. \ref{fig:loading-estimate}, bottom). Throughout, the robot center of mass stays tightly regulated in the mid-foot frame (Table~\ref{tab:com-shift}), moving by at most $32$\,mm horizontally and $7$\,mm in height, showing that the EKF estimates allow the WB-MPC to maintain balance objectives.

\begin{table}[tb]
\centering
\caption{Robot CoM in the $z$-up mid-foot frame over the loading run of Fig.~\ref{fig:loading-estimate}.}
\label{tab:com-shift}
\footnotesize
\begin{tabular}{@{}ccccc@{}}
\toprule
Dimension & Mean [m] & Min [m] & Max [m] & Spread [m] \\
\midrule
$x$ & $-0.042$ & $-0.059$ & $-0.032$ & $0.027$ \\
$y$ & $-0.023$ & $-0.045$ & $-0.013$ & $0.032$ \\
$z$ & $\phantom{-}0.862$ & $\phantom{-}0.859$ & $\phantom{-}0.865$ & $0.007$ \\
\bottomrule
\end{tabular}
\end{table}

\subsubsection{Teleoperated pick up of asymmetric load}

To test the behavior of teleoperation under load, we form a proxy task designed to emulate picking a load from the ground in a squatting, pitched forward posture. The VR operator first squats and pitches the robot via the joystick commands in Table~\ref{tab:vr-mapping}. After loading the right forearm with $4.5$\,kg ($10$\,lb), the robot returns to the home standing posture, where the VR operator commands circular window-washing motions.

\begin{figure}[tb]
    \centering
    \includegraphics[width=\linewidth,height=!]{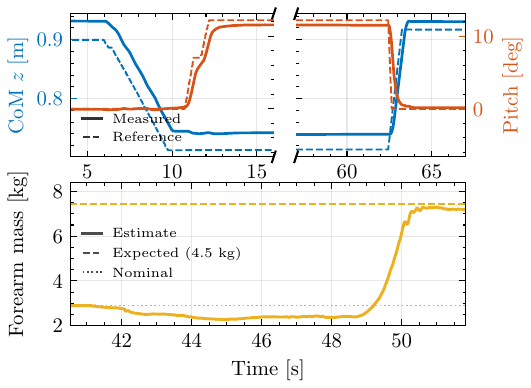}
    \caption{Top: Robot CoM height and pelvis pitch tracking through the squat/stand. Bottom: estimated forearm mass across the loading interval.}
    \label{fig:teleop-tracking}
\end{figure}

Tracking of the robot CoM height and pelvis pitch are generally good (Fig.~\ref{fig:teleop-tracking}, top) aside from a persistent $4$\,cm offset in the height, which we hypothesize is due to a cost conflict between the CoM tracking, lower body state regularization, and torque regularization in Table~\ref{tab:costs}. During the load interval, the estimate first undershoots the nominal mass due to manual handling of the link, but recovers $95\%$ of the added $4.5$\,kg load (Fig.~\ref{fig:teleop-tracking}, bottom).

\begin{figure}[tb]
    \centering
    \includegraphics[width=\linewidth,height=!]{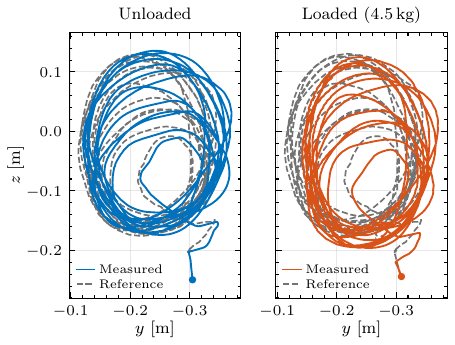}
    \caption{Taskspace paths (chest frame, frontal $y$–$z$ plane) for unloaded vs. loaded right forearm.}
    \label{fig:teleop-taskspace}
    \vspace{-6pt}
\end{figure}

To isolate the payload's effect on the loaded forearm, the arm streaming references commanded via VR to the WB-MPC while loaded were recorded and then replayed with the forearm unloaded, so both runs track the same commanded trajectory (Fig.~\ref{fig:teleop-taskspace}). As can be seen in Table~\ref{tab:teleop-tracking}, the loaded forearm sags $15.5$\,mm vertically from its reference and accumulates a vertical RMSE of $22.4$\,mm, versus $13.2$\,mm unloaded, which we consider a modest degradation given the $4.5$\,kg payload. The horizontal axis is far less affected: the payload increases the bias by only $4$\,mm and its RMSE by $5$\,mm, highlighting that under the benign conditions of teleoperation, the gravitational component of an added load dominates the tracking error.

\begin{table}[tb]
\centering
\caption{Right forearm taskspace error statistics (chest frame, frontal $y$--$z$ plane), loaded vs. unloaded baseline.}
\label{tab:teleop-tracking}
\footnotesize
\begin{tabular}{@{}lcccc@{}}
\toprule
 & \multicolumn{2}{c}{RMSE [mm]} & \multicolumn{2}{c}{Bias [mm]} \\
\cmidrule(lr){2-3}\cmidrule(l){4-5}
Condition & $y$ & $z$ & $y$ & $z$ \\
\midrule
Unloaded        & 14.2 & 13.2 & $-11.5$ & $\phantom{-}0.8$ \\
Loaded (4.5\,kg) & 19.3 & 22.4 & $-15.6$ & $-15.5$ \\
\bottomrule
\end{tabular}
\end{table}

\section{Conclusions and Future Work}

In this paper we showcased practical VR teleoperation of unknown payloads with a combination of whole body model predictive control and physically consistent inertial extended Kalman filtering on a humanoid robot.

Simulation experiments showed the necessity of online inertial estimation for trajectory tracking via the WB-MPC with significant added payload, and how closed-loop coupling of the EKF and WB-MPC allows an emergent adjustment of posture under load to maintain balance. The framework is further validated on hardware by an operator test of the VR interface; the robot squats down to pick up and manipulate a large asymmetric load.

In future work, we would like to incorporate perception techniques for producing inertial priors of objects that are due to be manipulated, to reduce the estimation transient response. We would also like to more thoroughly investigate the limits of robustness while performing more dynamic maneuvers with a grasped object. Finally, it would be interesting to integrate this framework with state-of-art large behavior models, whose typical output of taskspace poses can naturally be consumed by the WB-MPC controller.

\bibliographystyle{IEEEtran}
\bibliography{IEEEabrv, references}

\end{document}